# ParkSense: Where Should a Delivery Driver Park? Leveraging Idle AV Compute and Vision-Language Models

Die Hu[1]     Henan Li[2]

[1]University of California, Santa Barbara     [2]Independent Researcher
Correspondence: diehu@ucsb.edu

**Abstract.** Finding parking consumes a disproportionate share of food delivery time, yet no system addresses precise parking-spot selection relative to merchant entrances. We propose ParkSense, a framework that repurposes idle compute during low-risk AV states—queuing at red lights, traffic congestion, parking-lot crawl—to run a Vision-Language Model (VLM) on pre-cached satellite and street view imagery, identifying entrances and legal parking zones. We formalize the Delivery-Aware Precision Parking (DAPP) problem, show that a quantized 7B VLM completes inference in 4–8 seconds on HW4-class hardware, and estimate annual per-driver income gains of $3,000–$8,000 in the U.S. Five open research directions are identified at this unexplored intersection of autonomous driving, computer vision, and last-mile logistics.

**Keywords:** autonomous driving; delivery logistics; parking optimization; vision-language models; compute reallocation

## 1 Introduction

### 1.1 The Hidden Cost of Parking in Delivery

The global last-mile delivery market reached approximately $160 billion in 2024 (Precedence Research), accounting for an estimated 41–53% of total supply chain delivery costs [1]. Within last-mile operations, parking is a surprisingly dominant bottleneck. Butrina et al. found that the true constraint is the "last 800 feet"—from vehicle stop to merchant door [2]. Urban drivers spend a median of 8–10 minutes searching for parking in dense city centers [3], and approximately 60% of delivery drivers resort to illegal parking when legal spots are unavailable [4]. In New York City, delivery companies collectively paid $102 million in parking fines in 2006 [5]; UPS alone paid $23 million in 2019 [6], confirming the problem persists.

### 1.2 Motivation from Practice

This research originates from firsthand observations during food delivery work. A delivery driver approaching a merchant performs a rapid spatial reasoning task: *Where is the entrance—front, side, or rear? Which side of the street allows stopping? Is there a loading zone?* The workflow involves zooming into satellite view to read the building layout from above, then often switching to street view to confirm door locations and parking signage. At a walking speed of 1.2 m/s, each 10 m of closer parking saves approximately 8 seconds per trip; across 30–50 deliveries per day, these seconds compound into meaningful income differences—an observation consistent with findings by the Urban Freight Lab [2] and curb-availability studies [7]. This paper seeks to automate precisely this practitioner workflow.

### 1.3 Gaps in Current Solutions

No existing system addresses delivery-oriented precision parking:

- **Navigation applications** terminate at address-level coordinates (20–50 m GPS error in urban canyons), offering no entrance-level guidance.
- **Tesla FSD V14.1** introduced destination parking with five mode categories (Street, Parking Lot, Driveway, Parking Garage, Curbside) [8], selecting a *category* rather than a precise location relative to a specific entrance. FSD V14.2 showed mixed results in parking-lot scenarios [9].
- **Autonomous delivery platforms** (Nuro, Serve Robotics, Meituan) rely on pre-set delivery points or require the recipient to approach the vehicle [10].
- **Map platform APIs**: Google Places API does not include entrance coordinates. The Geocoding API offers entrance data via `extra_computations=BUILDING_AND_ENTRANCES` [11], but coverage is limited to large buildings. Apple MapKit provides no entrance data (as of early 2026).
- **Crowdsourced approaches**: Platforms like Uber use historical GPS traces for pickup-point optimization, and Muriel et al. [12] applied reinforcement learning to delivery parking route decisions. However, these approaches require substantial historical data per location and cannot handle new merchants or changed conditions.

### 1.4 Key Insight and Contribution

While Chen et al. [13] established the general concept of repurposing vehicle compute for non-driving services ("Vehicle as a Service"), and Xiao [14] demonstrated VLM-assisted last-meter delivery navigation for robots, no prior work has combined idle AV compute with VLM-based pre-journey imagery analysis for delivery parking optimization.

We observe that autonomous vehicles in urban delivery spend significant time in low-risk stationary or near-stationary states: queued behind other vehicles at red lights (where front cars provide a physical buffer), stuck in traffic congestion, or crawling through parking lots. In these states, trajectory prediction and motion planning are largely idle, and perception can operate at reduced throughput. We estimate 30–60% of compute becomes available depending on scenario (see Section 3.2), yielding 75–300 TOPS on current platforms —sufficient for a quantized 7B VLM. Critically, the driving stack is never unloaded; the VLM process is terminated within milliseconds when conditions change.

Our contributions:
1. **Problem definition**: We formalize Delivery-Aware Precision Parking (DAPP) as a monetized constrained optimization.
2. **Hybrid architecture**: ParkSense combines crowdsourced fleet data (primary) with VLM-based imagery analysis (complementary, for cold-start and long-tail scenarios).
3. **Feasibility demonstration**: Current AV hardware supports multiple VLM inference passes within a single red-light cycle.
4. **Economic analysis**: We quantify per-driver and platform-level impact for the U.S. market.
5. **Research roadmap**: Five open directions at this unexplored intersection.

## 2 Problem Formulation

### 2.1 Delivery-Aware Precision Parking (DAPP)

**Inputs:** Destination $D$; building entrances $E = \{e_1, ..., e_n\}$; candidate parking positions $P = \{p_1, ..., p_m\}$; time-dependent legality $L(p, t) \in \{\text{legal}, \text{time-limited}, \text{illegal}\}$; violation risk $R(p, t) \in [0, 1]$; fine amount $\text{Fine}(p)$; driver hourly wage $w$.

**Entrance inference** is a prerequisite sub-problem: given satellite imagery, street view, and map data for $D$, infer $E$. Separating this from the optimization ensures DAPP remains well-defined regardless of how $E$ is obtained.

**Candidate set $P$:** Generated by sampling at 5–10 m intervals along curbside lanes within 200 m of $D$, pre-filtered by structured parking data. Typical $|P| = 50$–200, trivially enumerable.

**Monetized objective** (all terms in $):

$$\min_{p \in P} \quad C_{\text{walk}} + C_{\text{park}} + C_{\text{risk}}$$

where:

$$e^* = \arg\min_{e \in E} \text{Walk}(p, e)$$
$$C_{\text{walk}} = \frac{\text{Walk}(p, e^*)}{v_{\text{walk}}} \times (w/3600)$$
$$C_{\text{park}} = \text{ParkTime}(p) \times (w/3600)$$
$$C_{\text{risk}} = R(p, t) \times \text{Fine}(p)$$

subject to $L(p, t) \in \{\text{legal}, \text{time-limited}\}$.

Walk$(p, e)$ denotes pedestrian network distance; when sidewalk data is unavailable, approximated as $1.4 \times$ Euclidean distance [15]. $v_{\text{walk}} \approx 1.2$ m/s. ParkTime ranges from 15 s (curbside pull-over) to 60 s (parallel parking).

**Fallback:** If $P_{\text{legal}} = \emptyset$, the system switches to soft-constraint mode ($R(p, t) < \tau$, e.g., $\tau = 0.1$), alerting the driver that only risk-tolerant options are available.

## 2.2 Violation Risk Function

$R(p, t)$ can be derived from: (a) structured parking-rule data (SpotAngels, OSM), yielding $R = 0$ for legal or $R = 1$ for clearly illegal; (b) historical enforcement data or parking-ticket datasets [16]; or (c) VLM detection of ambiguous signage, yielding intermediate values.

# 3 The ParkSense Framework

## 3.1 Hybrid Architecture: Crowdsourced + VLM

A key design decision is that **crowdsourced fleet data is the primary recommendation source** for frequently visited merchants (estimated 60–70% of deliveries). Delivery platforms accumulate millions of GPS traces, naturally forming optimal-parking heatmaps per merchant. ParkSense's VLM analysis serves as the **complementary layer** for:

- **Cold-start**: New merchants with no historical data.
- **Long-tail**: Merchants visited fewer than 10 times, where crowdsourced data is insufficient.
- **Dynamic changes**: New construction, changed entrances, temporary no-parking zones.

This hybrid design means VLM inference is triggered for approximately 30–40% of deliveries, substantially reducing compute and API demands.

### 3.2 Four-Phase Architecture

**Phase 1 — Pre-cache (navigation start):** Priority download: (1) satellite tiles ( 1 MB) + structured data (OSM, parking rules, <1 MB) in <5 s on 4G; (2) nearest street view images ( 5–10 MB); (3) extended coverage ( 20–80 MB, background). For frequently visited merchants, cached results eliminate fresh downloads.

**Phase 2 — Compute release (vehicle in low-risk state):** We identify a spectrum of scenarios where compute can be partially reallocated:

| Scenario | Risk | Available compute | Duration |
|---|---|---|---|
| Deep in queue (3rd+ car) | Very low | 45–60% | 30–120 s |
| Traffic congestion | Very low | 40–55% | Minutes |
| Front of queue at red | Low | 30–45% | 30–90 s |
| Stopped in parking lot | Very low | 50–65% | Variable |
| Waiting for left turn | Low–moderate | 25–40% | 15–45 s |

Table 1: Compute availability across low-risk driving scenarios. A vehicle queued behind others faces minimal lateral threat; the cars ahead serve as a physical buffer.

Critically, the driving stack is **never shut down**—it continues at reduced throughput. The VLM runs as a secondary process, **terminated within milliseconds** when conditions change. There is no model "reload" penalty; the driving stack simply reclaims freed GPU/NPU cycles.

**Phase 3 — VLM inference (on freed compute):** A quantized 7B VLM (e.g., Qwen2-VL-7B INT4) performs sequential analyses with structured JSON output: satellite pass (building layout, entrance inference), street view pass (entrance confirmation, sign detection), constraint overlay, and candidate ranking. Each pass takes approximately **4–8 seconds on HW4-class hardware**. A 45-second red light permits 5–6 passes.

**Phase 4 — Output (approaching destination):** Ranked parking recommendations with estimated walk distances. Real-time perception validates upon arrival.

### 3.3 Safety Considerations

Parking recommendations are QM-level (non-safety-critical) under ISO 26262 [17]. The VLM runs in a hardware-isolated partition via platform-specific domain isolation, fully preemptible by the ASIL-B to ASIL-D driving stack.

### 3.4 Data Fusion (U.S. Market)

| Layer | Source | Coverage |
|---|---|---|
| Entrances | Geocoding API + OSM + VLM | Partial; VLM fills gaps |
| Parking rules | SpotAngels + OSM | Major U.S. cities |
| Visual | Street View + satellite + VLM | Broad urban |
| Historical | Crowdsourced fleet GPS | High-frequency merchants |

Table 2: Multi-source data fusion layers for the U.S. market.

### 3.5 Deployment Model and Cost

Google Street View Static API costs $7/1,000 requests [18]. At 40–80 images per destination, per-delivery cost is $0.30–0.60—prohibitive for individual drivers. The viable model is **platform-level pre-processing**: delivery platforms cache imagery for active merchants, reducing marginal cost to <$0.01/delivery.

## 4 Feasibility Analysis

### 4.1 Compute Budget

| Platform | Capacity | Est. idle | 7B VLM? |
|---|---|---|---|
| Tesla HW3 | 144 TOPS INT8 [19] | 45–85 | 2–3B only |
| Tesla HW4 | est. 300–500 TOPS [20] | 90–300 | **Yes** |
| NVIDIA Orin | 254 TOPS INT8 [21] | 75–150 | Yes |
| NVIDIA Thor | 2,000 TOPS FP8 [21] | 600–1,200 | Easily |

Table 3: Compute budget across AV platforms. Idle estimates use 30–60% range (scenario-dependent). A 7B INT4 VLM requires ~60 TOPS.

### 4.2 Inference Timing

Published Jetson AGX Orin benchmarks (200 TOPS dense INT8 [22]) show 7B INT4 text models at 15–25 tokens/s, with VLM image encoding adding 2–4 s first-token latency. For one image + 150-token output: **~6–12 s on Orin-class, ~4–8 s on HW4-class** (extrapolated). These estimates require vehicle-grade validation.

### 4.3 Red-Light Time Budget

U.S. urban signals operate on 60–120 s cycles, with per-phase red times of ~25–90 s [23]. At a conservative 30 s and 8 s per pass: 3 passes. At 60 s: 7 passes. Multiple en-route red lights extend the budget further.

## 5 Economic Analysis

### 5.1 Per-Driver Impact (U.S.)

U.S. gig delivery drivers earn $15–25/hour, completing 20–40 deliveries per day. Studies indicate parking search time of 5–8 minutes in dense urban areas [3], of which we estimate ParkSense can reduce 30–50%—yielding **1.5–4 minutes saved per delivery**.

*Derivation*: At $20/hour, 1 min = $0.33. Saving 2.5 min × 30 deliveries = $25/day. Over 250 working days: **~$6,000/year**. Conservative: ~$3,000/year. Optimistic: ~$8,000/year. This assumes saved time translates to additional deliveries given sufficient order supply.

### 5.2 Platform-Level Impact

With several million U.S. gig delivery workers [24], fleet-wide savings of 3 minutes daily free millions of driver-minutes—equivalent to tens of thousands of additional daily deliveries.

## 5.3 Violation Cost Reduction

NYC delivery firms paid $102 million in fines (2006) [5]; UPS alone paid $23 million (2019) [6]. ParkSense's legal-spot recommendations could reduce individual driver fines by hundreds of dollars annually.

## 6 Limitations

Several core assumptions require empirical validation: (1) the scenario-dependent idle compute estimates (30–60%) have not been measured on production AV hardware; (2) VLM accuracy for entrance detection from satellite imagery is scenario-dependent—likely reliable for standalone buildings but unreliable for row shops, underground entrances, or canopy occlusion; (3) street view and satellite imagery may be outdated; (4) the economic model assumes saved time converts to additional deliveries. The VLM approach is complementary to, not a replacement for, crowdsourced data—which remains more reliable for high-frequency merchants.

## 7 Research Roadmap

1. **Scenario-adaptive compute reallocation.** Building on VaaS [13], formal models for partitioning AV compute between driving and auxiliary tasks, validated with empirical measurements.
2. **VLM geospatial parking reasoning benchmarks.** Quantifying VLM accuracy on entrance-side detection and sign recognition—extending [14] to pre-journey cached imagery.
3. **Pre-journey destination scene understanding.** Cached remote imagery analyzed *before arrival*, complementing real-time perception.
4. **Delivery-aware parking benchmarks.** We propose *DeliveryPark-US*, covering NYC, SF, LA, Chicago, and Houston.
5. **Parking regulation digitization.** Unifying fragmented curbside-rule data into a queryable geospatial layer.

## 8 Conclusion

This paper identifies delivery parking optimization as a significant, quantifiable, and unsolved problem. ParkSense demonstrates that idle AV compute at traffic stops can power VLM-based destination pre-analysis —technically feasible on today's hardware. Combined with crowdsourced fleet data as the primary layer, VLM analysis fills critical gaps for new, infrequent, and dynamically changing destinations. The DAPP formalization and hybrid architecture provide a foundation for advancing precision parking from category-level to entrance-level.

**Acknowledgments.** This research was motivated by practical experience in food delivery, where parking decisions consistently dominated non-driving time. The views expressed are those of the authors. No university resources were used for this work.